# A Generalized Loop Correction Method for Approximate Inference in Graphical Models

**Siamak Ravanbakhsh** MRAVANBA@UALBERTA.CA
**Chun-Nam Yu** CHUNNAM@UALBERTA.CA
**Russell Greiner** RGREINER@UALBERTA.CA
Department of Computing Science, University of Alberta, Edmonton, AB T6G 2E8 CANADA

## Abstract

Belief Propagation (BP) is one of the most popular methods for inference in probabilistic graphical models. BP is guaranteed to return the correct answer for tree structures, but can be incorrect or non-convergent for loopy graphical models. Recently, several new approximate inference algorithms based on *cavity distribution* have been proposed. These methods can account for the effect of loops by incorporating the dependency between BP messages. Alternatively, *region-based* approximations (that lead to methods such as Generalized Belief Propagation) improve upon BP by considering interactions within small clusters of variables, thus taking small loops within these clusters into account. This paper introduces an approach, *Generalized Loop Correction* (GLC), that benefits from both of these types of loop correction. We show how GLC relates to these two families of inference methods, then provide empirical evidence that GLC works effectively in general, and can be significantly more accurate than both correction schemes.

## 1. Introduction

Many real-world applications require probabilistic inference from some known probabilistic model (Koller & Friedman, 2009). This paper will use probabilistic graphical models, focusing on factor graphs (Kschischang et al., 1998), that can represent both Markov Networks and Bayesian Networks. The basic challenge of such inference is marginalization (or max-marginalization) over a large number of variables. For discrete variables, computing the exact solutions is NP-hard, typically involving a computation that is exponential in the number of variables.

When the conditional dependencies of the variables form a tree structure (*i.e.*, no loops), this exact inference is tractable, and can be done by a message passing procedure, Belief Propagation (BP) (Pearl, 1988). The Loopy Belief Propagation (LBP) system applies BP repeatedly to graph structures that are not trees (called "loopy graphs"); however, this provides only an approximately correct solution (when it converges).

LBP is related to the Bethe approximation to free energy (Heskes, 2003), which is the basis for minimization of more sophisticated energy approximations and provably convergent methods (Yedidia et al., 2005; Heskes, 2006; Yuille, 2002). A representative class of energy approximations is the region-graph methods (Yedidia et al., 2005), which deal with a set of connected variables (called "regions"); these methods subsume both the Cluster Variation Method (CVM) (Pelizzola, 2005; Kikuchi, 1951) and the Junction Graph Method (Aji & McEliece, 2001). Such region-based methods deal with the short loops of the graph by incorporating them into overlapping regions (see Figure 1(a)), and perform exact inference over each region. Note a valid region-based methods is exact if its *region graph* has no loops.

A different class of algorithms, *loop correction methods*, tackles the problem of inference in loopy graphical models by considering the *cavity distribution* of variables. A *cavity distribution* is defined as the marginal distribution on Markov blanket of a single (or a cluster of) variable(s), after removing all factors that depend on those initial variables. Figure 1(b) illustrates cavity distribution, and also shows that the cavity variables can interact. The key observation in these methods is that, by removing a variable $x_i$ in a graphical model, we break all the loops that involve the variable $x_i$, resulting in a simplified problem of finding the cavity distribution. The marginals around $x_i$ can then be recovered by considering the cavity distribution and its interaction with $x_i$. This is the basis for the loop correction schemes by Montanari & Rizzo's (2005) on pairwise dependencies over binary variables, and also Mooij & Kappen's (2007) extension to general factor graphs – called Loop Corrected Belief Propagation (LCBP).

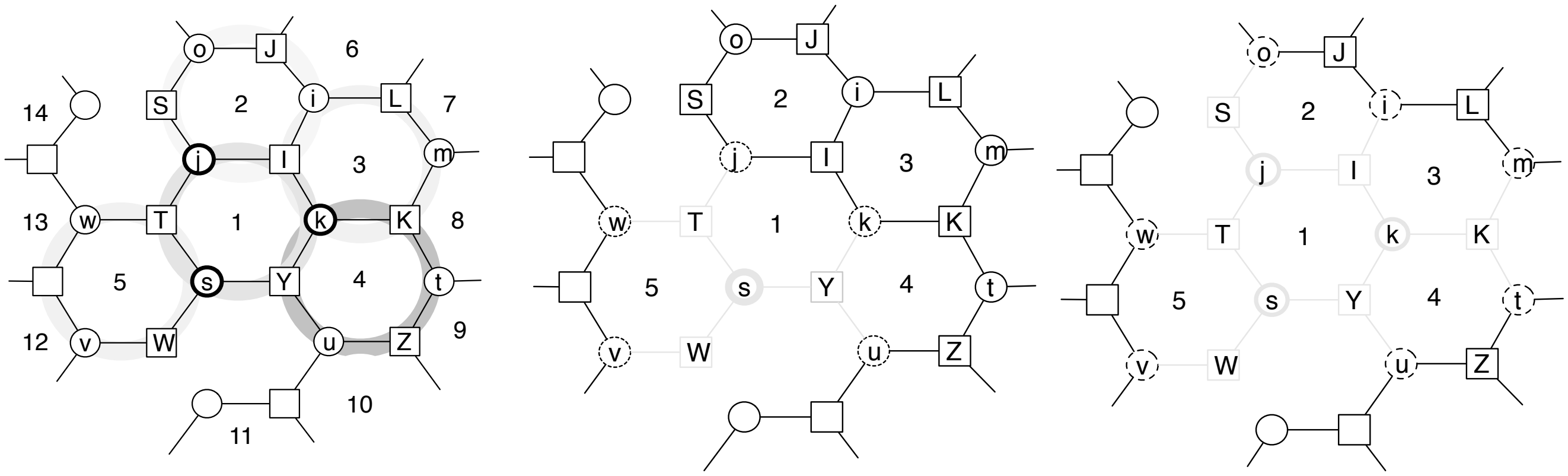

*Figure 1.* Part of a factor graph, where circles are variables (circle labeled "$i$" corresponding to variable "$\boldsymbol{x}_i$") and squares (with CAPITAL letters) represent factors. Note variables $\{\boldsymbol{x}_i, \boldsymbol{x}_k, \boldsymbol{x}_s\}$ form a loop, as do $\{\boldsymbol{x}_k, \boldsymbol{x}_u, \boldsymbol{x}_t\}$, etc.
(a) An example of absorbing short loops into overlapping regions. Here, a region includes factors around each hexagon and all its variables. Factor $I$ and the variables $\boldsymbol{x}_i, \boldsymbol{x}_j, \boldsymbol{x}_k$ appear in the three regions $\boldsymbol{r}_1$, $\boldsymbol{r}_2$, $\boldsymbol{r}_3$. (Figure just shows index $\alpha$ for region $\boldsymbol{r}_\alpha$.) Region-based methods provide a way to perform inference on overlapping regions. (In general, regions do not have to involve exactly 3 variables and 3 factors.)
(b) Cavity variables for $\boldsymbol{x}_s$ are $\{\boldsymbol{x}_w,\ \boldsymbol{x}_j,\ \boldsymbol{x}_k,\ \boldsymbol{x}_u,\ \boldsymbol{x}_v\}$, shown using dotted circles. We define the cavity distribution for $\boldsymbol{x}_s$ by removing all the factors around this variable, and marginalizing the remaining factor-graph on dotted circles. Even after removing factors $\{T, Y, W\}$, the variables $\boldsymbol{x}_v, \boldsymbol{x}_w$, and $\boldsymbol{x}_j, \boldsymbol{x}_k, \boldsymbol{x}_u$ still have higher-order interactions caused by remaining factors, due to loops in the factor graph.
(c) Cavity region $\boldsymbol{r}_1 = \{j, s, k\}$ includes variables shown in pale circles. Variables in dotted circles are the perimeter $\ominus\boldsymbol{r}_1$. Removing the "pale factors" and marginalizing the rest of network on $\ominus\boldsymbol{r}_1$, gives the cavity distribution for $\boldsymbol{r}_1$.

This paper defines a new algorithm for probabilistic inference, called *Generalized Loop Correction* (GLC), that uses a more general form of cavity distribution, defined over *region*s, and also a novel message passing scheme between these regions that uses cavity distributions to correct the types of loops that result from exact inference over each region. GLC's combination of loop corrections is well motivated, as region-based methods can deal effectively with short loops in the graph, and the approximate cavity distribution is known to produce superior results when dealing with long influencial loops (Mooij & Kappen, 2007).

In its simplest form, GLC produces update equations similar to LCBP's; indeed, under a mild assumption, GLC reduces to LCBP for pairwise factors. In its general form, when not provided with information on cavity variable interactions, GLC produces results similar to region-based methods. We theoretically establish the relation between GLC and region-based approximations, for a limited setting.

Section 2 explains the notation, factor graph representation and preliminaries for GLC. Section 3 introduces a simple version of GLC that works with regions that partition the set of variables; followed by its extension to the more general algorithm. Section 4 presents empirical results, comparing our GLC against other approaches.

## 2. Framework

### 2.1. Notation

Let $\boldsymbol{X} = \{X_1, X_2, \ldots, X_N\}$ be a set of $N$ discrete-valued random variables, where $X_i \in \mathcal{X}_i$. Suppose their joint probability distribution factorizes into a product of non-negative functions:

$$P(\boldsymbol{X} = \boldsymbol{x}) \quad := \quad \frac{1}{Z} \prod_{I \in \mathcal{F}} \psi_I(\boldsymbol{x}_I)$$

where each $I \subseteq \{1, 2, \ldots, N\}$ is a subset of the variable indices, and $\boldsymbol{x}_I = \{x_i \mid i \in I\}$ is the set of values in $\boldsymbol{x}$ indexed by the subset $I$. Each factor $\psi_I : \prod_{i \in I} \mathcal{X}_i \to [0, \infty)$ is a non-negative function, and $\mathcal{F}$ is the collection of indexing subsets $I$ for all the factors $\psi_I$. Below we will use the term "factor" interchangeably with the function $\psi_I$ and subset $I$, and the term "variable" interchangeably for the value $x_i$ and index $i$. Here $Z$ is the partition function.

This model can be conveniently represented as a bipartite graph, called the *factor graph* (Kschischang et al., 1998), which includes two sets of nodes: variable nodes $x_i$, and factor nodes $\psi_I$. A variable node $x_i$ is connected to a factor node $\psi_I$ if and only if $i \in I$. We use the notation $N(i)$ to denote the neighbors of variable $x_i$ in the factor graph – *i.e.*, the set of factors defined by $N(i) := \{I \in \mathcal{F} \mid i \in I\}$. To illustrate, using Figure 1(a): $N(j) = \{I, T, S\}$ and $T = \{j, s, w\}$.

We use the shorthand $\psi_{\mathcal{A}}(\boldsymbol{x}) := \prod_{I \in \mathcal{A}}(\boldsymbol{x}_I)$ to denote the product of factors in a set of factors $\mathcal{A}$. For marginalizing all possible values of $\boldsymbol{x}$ except the $i$th variable, we define the notation:

$$\sum_{\boldsymbol{x}\setminus i} f(\boldsymbol{x}) \quad := \quad \sum_{x_j \in \mathcal{X}_j, j \neq i} f(\boldsymbol{x}).$$

Similarly for a set of variables $\boldsymbol{r}$, we use the notation $\sum_{\boldsymbol{x}\setminus \boldsymbol{r}}$ to denote marginalization of all variables apart from those in $\boldsymbol{r}$.

### 2.2. Generalized Cavity Distribution

The notion of *cavity distribution* is borrowed from so-called *cavity methods* from statistical physics (Mézard & Montanari, 2009), and has been used in analysis and optimization of important combinatorial problems (Mézard et al., 2002; Braunstein et al., 2002). The basic idea is to make a cavity by removing a variable $\boldsymbol{x}_i$ along with all the factors around it, from the factor graph (Figure 1(b)). We will use a more general notion of regional cavity, around a region.

**Definition** A *cavity region* is a subset of variables $\boldsymbol{r} \subseteq \{1, \ldots, N\}$ that are connected by a set of factors – *i.e.*, the set of variable nodes $\boldsymbol{r}$ and the associated factors $N(\boldsymbol{r}) := \{N(i) \mid i \in \boldsymbol{r}\}$ forms a connected component on the factor graph.

For example in Figure 1(a), the variables indexed by $\boldsymbol{r}_1 = \{j, k, s\}$ define a cavity region with factors $N(\boldsymbol{r}_1) = \{I, T, Y, S, W, K\}$

**Remark** A "cavity region" is different from common notion of region in region-graph methods, in that a cavity region includes all factors in $N(\boldsymbol{r})$ (and nothing more), while common regions allow a factor $I$ to be a part of a region only if $I \subseteq \boldsymbol{r}$.

The notation $\oplus\boldsymbol{r} := \{i \in I \mid I \in N(\boldsymbol{r})\}$ denotes the cavity region $\boldsymbol{r}$ with its surrounding variables, and $\ominus\boldsymbol{r} := \oplus\boldsymbol{r} \setminus \boldsymbol{r}$ denotes just the perimeter of the cavity region $\boldsymbol{r}$. In Figure 1(c), the dotted circles show the indices $\ominus\boldsymbol{r}_1 = \{o, i, m, t, u, v, w\}$ and their union with the pale circles defines $\oplus\boldsymbol{r}_1$.

**Definition** The *Cavity Distribution*, for cavity region $\boldsymbol{r}$, is defined over the variables indexed by $\ominus\boldsymbol{r}$, as:

$$P^{\setminus \boldsymbol{r}}(\boldsymbol{x}_{\ominus\boldsymbol{r}}) \propto \sum_{\boldsymbol{x}_{\setminus\ominus\boldsymbol{r}}} \psi_{\mathcal{F}\setminus N(\boldsymbol{r})}(\boldsymbol{x}) = \sum_{\boldsymbol{x}_{\setminus\ominus\boldsymbol{r}}} \prod_{I \notin N(\boldsymbol{r})} \psi_I(\boldsymbol{x}_I)$$

Here the summation is over all variables but the ones indexed by $\ominus\boldsymbol{r}$.

In Figure 1(c), this is the distribution obtained by removing factors $N(\boldsymbol{r}_1) = \{I, T, Y, K, S, W\}$ from the factor gaph and marginalizing the rest over dotted circles, $\ominus\boldsymbol{r}_1$.

The core idea to our approach is that each cavity region $\boldsymbol{r}$ can produce reliable probability distribution over $\boldsymbol{r}$, given an accurate cavity distribution estimate over the surrounding variables $\ominus\boldsymbol{r}$. Given the exact cavity distribution $P^{\setminus \boldsymbol{r}}$ over $\ominus\boldsymbol{r}$, we can recover the exact joint distribution $P_{\boldsymbol{r}}$ over $\oplus\boldsymbol{r}$ by:

$$P_{\boldsymbol{r}}(\boldsymbol{x}_{\oplus\boldsymbol{r}}) \propto P^{\setminus \boldsymbol{r}}(\boldsymbol{x}_{\ominus\boldsymbol{r}})\psi_{N(\boldsymbol{r})}(\boldsymbol{x}) = P^{\setminus \boldsymbol{r}}(\boldsymbol{x}_{\ominus\boldsymbol{r}}) \prod_{I \in N(\boldsymbol{r})} \psi_I(\boldsymbol{x}_I) \,.$$

In practice, we can only obtain estimates $\hat{P}^{\setminus \boldsymbol{r}}(\boldsymbol{x}_{\ominus\boldsymbol{r}})$ of the true cavity distribution $P^{\setminus \boldsymbol{r}}(\boldsymbol{x}_{\ominus\boldsymbol{r}})$. However, suppose we have multiple cavity regions $\boldsymbol{r}_1, \boldsymbol{r}_2, \ldots, \boldsymbol{r}_M$ that collectively cover all the variables $\{x_1, \ldots, x_N\}$. If $\ominus\boldsymbol{r}_p$ intersects with $\boldsymbol{r}_q$, we can improve the estimate of $\hat{P}^{\setminus \boldsymbol{r}_p}(\boldsymbol{x}_{\ominus\boldsymbol{r}_p})$ by enforcing marginal consistency of $\hat{P}_{\boldsymbol{r}_p}(\boldsymbol{x}_{\oplus\boldsymbol{r}_p})$ with $\hat{P}_{\boldsymbol{r}_q}(\boldsymbol{x}_{\oplus\boldsymbol{r}_q})$ over the variables in their intersection. This suggests an iterative correction scheme that is very similar to message passing.

In Figure 1(a), let each hexagon (over variables and factors) define a cavity region, here $\boldsymbol{r}_1, \ldots, \boldsymbol{r}_5$. Note $\boldsymbol{r}_1$ can provide good estimates over $\{j, s, k\}$, given good approximation to cavity distribution over $\{o, i, m, t, u, v, w\}$. This in turn can be improved by neighboring regions; *e.g.*, $\boldsymbol{r}_2$ gives a good approximation over $\{i, o\}$, and $\boldsymbol{r}_3$ over $\{i, m\}$. Starting from an initial cavity distribution $\hat{P}_0^{\setminus \boldsymbol{r}_\alpha}$, for each cavity region $\alpha \in \{1, \ldots, 14\}$, We perform this improvement for all cavity regions, in iterations until convergence.

When we start with a uniform cavity distribution $\hat{P}_0^{\setminus \boldsymbol{r}_p}$ for all regions, the results are very similar to those of CVM. The accuracy of this approximation depends on the accuracy of the initial $\hat{P}_0^{\setminus \boldsymbol{r}_p}$.

Following Mooij (2008), we use variable clamping to estimate higher-order interactions in $\ominus\boldsymbol{r}$: Here, we estimate the partition function $Z_{\boldsymbol{x}_{\ominus\boldsymbol{r}}}$ after removing factors in $N(\boldsymbol{r})$ and fixing $\boldsymbol{x}_{\ominus\boldsymbol{r}}$ to each possible assignment. Doing this calculation, we have $\hat{P}^{\setminus \boldsymbol{r}}(\boldsymbol{x}_{\ominus\boldsymbol{r}}) \propto Z_{\boldsymbol{x}_{\ominus\boldsymbol{r}}}$. In our experiments, we use the approximation to the partition function provided using LBP. However there are some alternatives to clamping: conditioning scheme Rizzo et al. (2007) makes it possible to use any method capable of *marginalization* for estimation of cavity distribution (clamping requires estimation of *partition function*). It is also possible to use techniques in answering *joint queries* for this purpose (Koller & Friedman (2009)).

Using clamping for this purpose also means that, if the resulting network, after clamping, has no loops, then $\hat{P}_{\boldsymbol{r}}(\boldsymbol{x}_{\oplus \boldsymbol{r}})$ is exact – hence GLC produces exact results if for every cluster $\boldsymbol{r}$, removing $\oplus \boldsymbol{r}$ results in a tree.

## 3. Generalized Loop Correction

### 3.1. Simple Case: Partitioning Cavity Regions

To introduce our approach, first consider a simpler case where the cavity regions $\boldsymbol{r}_1, \ldots, \boldsymbol{r}_M$ form a (disjoint and exhaustive) partition of the variables $\{1, \ldots, N\}$.

Let $\ominus \boldsymbol{r}_{p,q} := (\ominus \boldsymbol{r}_p) \cap \boldsymbol{r}_q$ denote the intersection of the perimeter $\ominus \boldsymbol{r}_p$ of $\boldsymbol{r}_p$ with another cavity region $\boldsymbol{r}_q$. (Note $\ominus \boldsymbol{r}_{p,q} \neq \ominus \boldsymbol{r}_{q,p}$). As $\boldsymbol{r}_1, \ldots, \boldsymbol{r}_M$ is a partition, each perimeter $\ominus \boldsymbol{r}_p$ is a disjoint union of $\ominus \boldsymbol{r}_{p,q}$ for $q = 1 \ldots M$ (some of which might be empty if $\boldsymbol{r}_p$ and $\boldsymbol{r}_q$ are not neighbors). Let $Nb(p)$ denote the set of regions $q$ with $\ominus r_{p,q} \neq \emptyset$. We now consider how to improve the cavity distribution estimate over $\ominus \boldsymbol{r}_p$ through update messages sent to each of the $\ominus \boldsymbol{r}_{p,q}$.

In Figure 1(a), the regions $\boldsymbol{r}_2, \boldsymbol{r}_4, \boldsymbol{r}_5, \boldsymbol{r}_7, \boldsymbol{r}_{11}, \boldsymbol{r}_{14}$ form a partitioning. Here, $\boldsymbol{r}_2$ with $\{m, k, s, w\} \subset \ominus \boldsymbol{r}_2$, receives updates over $\ominus \boldsymbol{r}_{2,7} = \{m\}$ from $\boldsymbol{r}_7$ and updates over $\ominus \boldsymbol{r}_{2,4} = \{k\}$ from $\boldsymbol{r}_4$. This last update ensures $\sum_{\boldsymbol{x} \setminus \{k\}} \hat{P}_{\boldsymbol{r}_2}(\boldsymbol{x}_{\oplus \boldsymbol{r}_2}) = \sum_{\boldsymbol{x} \setminus \{k\}} \hat{P}_{\boldsymbol{r}_4}(\boldsymbol{x}_{\oplus \boldsymbol{r}_4})$. Towards enforcing this equality, we introduce a message $m_{4 \to 2}(\boldsymbol{x}_{\ominus \boldsymbol{r}_{2,4}})$ into distribution over $\oplus \boldsymbol{r}_2$.

Here, the distribution over $\oplus \boldsymbol{r}_p$ becomes: $\hat{P}_{\boldsymbol{r}_p}(\boldsymbol{x}_{\oplus \boldsymbol{r}_p}) \propto$

$$\hat{P}_0^{\setminus \boldsymbol{r}_p}(\boldsymbol{x}_{\ominus \boldsymbol{r}_p}) \psi_{N(\boldsymbol{r}_p)}(\boldsymbol{x}_{\oplus \boldsymbol{r}_p}) \prod_{q \in Nb(p)} m_{q \to p}(\boldsymbol{x}_{\ominus \boldsymbol{r}_{p,q}}), \quad (1)$$

where $\hat{P}_{\boldsymbol{r}_p}$ denotes our estimate of the true distribution $P_{\boldsymbol{r}_p}$.

The messages $m_{q \to p}$ can be recovered by considering marginalization constraints. When $\boldsymbol{r}_p$ and $\boldsymbol{r}_q$ are neighbors, their distributions $\hat{P}_{\boldsymbol{r}_p}(\boldsymbol{x}_{\oplus \boldsymbol{r}_p})$ and $\hat{P}_{\boldsymbol{r}_q}(\boldsymbol{x}_{\oplus \boldsymbol{r}_q})$ should satisfy

$$\sum_{\boldsymbol{x} \setminus \oplus \boldsymbol{r}_p \cap \oplus \boldsymbol{r}_q} \hat{P}_{\boldsymbol{r}_p}(\boldsymbol{x}_{\oplus \boldsymbol{r}_p}) = \sum_{\boldsymbol{x} \setminus \oplus \boldsymbol{r}_p \cap \oplus \boldsymbol{r}_q} \hat{P}_{\boldsymbol{r}_q}(\boldsymbol{x}_{\oplus \boldsymbol{r}_q}).$$

We can divide both sides by the factor product $\psi_{N(\boldsymbol{r}_p) \cap N(\boldsymbol{r}_q)}(\boldsymbol{x})$, as the domain of the factors in $N(\boldsymbol{r}_p) \cap N(\boldsymbol{r}_q)$ is completely contained in $\oplus \boldsymbol{r}_p \cap \oplus \boldsymbol{r}_q$ and independent of the summation. Hence we have

$$\sum_{\boldsymbol{x} \setminus \oplus \boldsymbol{r}_p \cap \oplus \boldsymbol{r}_q} \frac{\hat{P}_{\boldsymbol{r}_p}(\boldsymbol{x}_{\oplus \boldsymbol{r}_p})}{\psi_{N(\boldsymbol{r}_p) \cap N(\boldsymbol{r}_q)}(\boldsymbol{x})} = \sum_{\boldsymbol{x} \setminus \oplus \boldsymbol{r}_p \cap \oplus \boldsymbol{r}_q} \frac{\hat{P}_{\boldsymbol{r}_q}(\boldsymbol{x}_{\oplus \boldsymbol{r}_q})}{\psi_{N(\boldsymbol{r}_p) \cap N(\boldsymbol{r}_q)}(\boldsymbol{x})}$$

As $\ominus \boldsymbol{r}_{p,q} \subset \oplus \boldsymbol{r}_p \cap \oplus \boldsymbol{r}_q$, this implies the weaker consistency condition:

$$\sum_{\boldsymbol{x} \setminus \ominus \boldsymbol{r}_{p,q}} \hat{P}_{\boldsymbol{r}_p}(\boldsymbol{x}_{\oplus \boldsymbol{r}_p}) \psi_{N(\boldsymbol{r}_p) \cap N(\boldsymbol{r}_q)}(\boldsymbol{x})^{-1} = \sum_{\boldsymbol{x} \setminus \ominus \boldsymbol{r}_{p,q}} \hat{P}_{\boldsymbol{r}_q}(\boldsymbol{x}_{\oplus \boldsymbol{r}_q}) \psi_{N(\boldsymbol{r}_p) \cap N(\boldsymbol{r}_q)}(\boldsymbol{x})^{-1}, \quad (2)$$

which we can use to derive update equations for $m_{q \to p}$.

Starting from the LHS of Eqn (2),

$$\begin{aligned} &\sum_{\boldsymbol{x} \setminus \ominus \boldsymbol{r}_{p,q}} \hat{P}_{\boldsymbol{r}_p}(\boldsymbol{x}_{\oplus \boldsymbol{r}_p}) \psi_{N(\boldsymbol{r}_p) \cap N(\boldsymbol{r}_q)}(\boldsymbol{x})^{-1} \\ \propto &\sum_{\boldsymbol{x} \setminus \ominus \boldsymbol{r}_{p,q}} \hat{P}_0^{\setminus \boldsymbol{r}_p}(\boldsymbol{x}_{\ominus \boldsymbol{r}_p}) \psi_{N(\boldsymbol{r}_p) \setminus N(\boldsymbol{r}_q)}(\boldsymbol{x}) \prod_{q' \in Nb(p)} m_{q' \to p}(\boldsymbol{x}_{\ominus \boldsymbol{r}_{p,q'}}) \\ \propto &\; m_{q \to p}(\boldsymbol{x}_{\ominus \boldsymbol{r}_{p,q}}) \sum_{\boldsymbol{x} \setminus \ominus \boldsymbol{r}_{p,q}} \hat{P}_0^{\setminus \boldsymbol{r}_p}(\boldsymbol{x}_{\ominus \boldsymbol{r}_p}) \psi_{N(\boldsymbol{r}_p) \setminus N(\boldsymbol{r}_q)}(\boldsymbol{x}) \prod_{\substack{q' \in Nb(p) \\ q' \neq q}} m_{q' \to p}(\boldsymbol{x}_{\ominus \boldsymbol{r}_{p,q'}}). \end{aligned}$$

Setting this proportional to the RHS of Eqn (2), we have the update equation

$$\begin{aligned} &m^{new}_{q \to p}(\boldsymbol{x}_{\ominus \boldsymbol{r}_{p,q}}) \\ &\propto \frac{\sum_{\boldsymbol{x} \setminus \ominus \boldsymbol{r}_{p,q}} \hat{P}_{\boldsymbol{r}_q}(\boldsymbol{x}_{\oplus \boldsymbol{r}_q}) \psi_{N(\boldsymbol{r}_p) \cap N(\boldsymbol{r}_q)}(\boldsymbol{x})^{-1}}{\sum_{\boldsymbol{x} \setminus \ominus \boldsymbol{r}_{p,q}} \hat{P}_0^{\setminus \boldsymbol{r}_p}(\boldsymbol{x}_{\ominus \boldsymbol{r}_p}) \psi_{N(\boldsymbol{r}_p) \setminus N(\boldsymbol{r}_q)}(\boldsymbol{x}) \prod_{\substack{q' \in Nb(p) \\ q' \neq q}} m_{q \to p}(\boldsymbol{x}_{\ominus \boldsymbol{r}_{p,q'}})} \\ &\propto \frac{\sum_{\boldsymbol{x} \setminus \ominus \boldsymbol{r}_{p,q}} \hat{P}_{\boldsymbol{r}_q}(\boldsymbol{x}_{\oplus \boldsymbol{r}_q}) \psi_{N(\boldsymbol{r}_p) \cap N(\boldsymbol{r}_q)}(\boldsymbol{x})^{-1}}{\sum_{\boldsymbol{x} \setminus \ominus \boldsymbol{r}_{p,q}} \hat{P}_{\boldsymbol{r}_p}(\boldsymbol{x}_{\oplus \boldsymbol{r}_p}) \psi_{N(\boldsymbol{r}_p) \cap N(\boldsymbol{r}_q)}(\boldsymbol{x})^{-1}} m_{q \to p}(\boldsymbol{x}_{\ominus \boldsymbol{r}_{p,q}}) \quad (3) \end{aligned}$$

The last line follows from multiplying the numerator and denominator by the current version of the message $m_{q \to p}$. At convergence, when $m_{q \to p}$ equals $m^{new}_{q \to p}$, the consistency constraints are satisfied. By repeating this update in any order, after convergence, the $\hat{P}_{\boldsymbol{r}}(\boldsymbol{x}_{\oplus \boldsymbol{r}})$s represent approximate marginals over each region.

The following theorem stablishes the relation between GLC and CVM in a limited setting.

**Theorem 1** *If the cavity regions partition the variables and all the factors involve no more than 2 variables, then any GBP fixed point of a particular CVM construction (details in Appendix A) is also a fixed point for* GLC*, starting from uniform cavity distributions* $\hat{P}_0^{\setminus \boldsymbol{r}} = 1$. *(Proof in Appendix A.)*

**Corollary 1** *If the factors have size two and there are no loops of size 4 in the factor graph, for single variable partitioning with uniform cavity distribution, any fixed points of LBP can be mapped to fixed points of GLC.*

**Proof** If there are no loops of size 4 then no two factors have identical domain. Thus the factors are all *maximal* and GBP applied to CVM with maximal factor domains, is the same as LBP. On the other hand (refering to CVM construction of Appendix A) under the given condition, GLC with single variable partitioning shares the fixed points of GBP applied to CVM with maximal factors. Therefore GLC shares the fixed points of LBP.

**Theorem 2** *If all factors have size two and no two factors have the same domain,* GLC *is identical to LCBP under single variable partitioning.*

**Proof** Follows from comparison of two update equations – *i.e.*, Eqn (3) and Eqn (5) in (Mooij & Kappen, 2007)– under the assumptions of the theorem.

### 3.2. General Cavity Regions

When cavity regions do not partition the set of variables, the updates are more involved. As the perimeter $\ominus\boldsymbol{r}_p$ is no longer partitioned, the $\ominus\boldsymbol{r}_{p,q}$'s are no longer disjoint.

For example in Figure 1, for $\boldsymbol{r}_1$ we have $\ominus\boldsymbol{r}_{1,2}=\{o,i\}$, $\ominus\boldsymbol{r}_{1,3}=\{i,m\}$, $\ominus\boldsymbol{r}_{1,4}=\{t,u\}$, $\ominus\boldsymbol{r}_{1,5}=\{v,w\}$ and also $\ominus\boldsymbol{r}_{1,6}=\{i\}$, $\ominus\boldsymbol{r}_{1,7}=\{m\}$, $\ominus\boldsymbol{r}_{1,8}=\{m,t\}$, $\ominus\boldsymbol{r}_{1,9}=\{t\}$, etc. This means $\boldsymbol{x}_i$ appears in messages $m_{2\to1}$, $m_{3\to1}$ and $m_{6\to1}$.

Directly adopting the correction formula for $\hat{P}_{\boldsymbol{r}}$ in Eqn (1) as a product of messsages over $\ominus\boldsymbol{r}_{p,q}$ could double-count variables. To avoid this problem, we adopt a strategy similar to CVM to discount extra contributions from overlapping variables in $\ominus\boldsymbol{r}_p$. For each cavity region $\boldsymbol{r}_p$, we form a $\ominus\boldsymbol{r}_p$-region graph (Figure 2) with the incoming messages forming the distributions over top regions. For computational reasons, we only consider maximal $\ominus\boldsymbol{r}_{p,q}$ domains.[1] here, this means dropping $m_{6\to1}$ as $\ominus\boldsymbol{r}_{1,6}\subset\ominus\boldsymbol{r}_{1,2}$ and so on.

Our region-graph construction is similar to CVM (Pelizzola, 2005) – *i.e.*, we construct new sub-regions as the intersection of $\ominus\boldsymbol{r}_{p,q}$'s, and we repeat this recursively until no new region can be added. We then connect each sub-region to its immediate parent. Figure 2 shows the $\ominus\boldsymbol{r}_1$-region graph for the example of Figure 1(a). If the cavity regions are a *partition*, the $\ominus\boldsymbol{r}_p$-region graph includes only the top regions. Below we use $\ominus\mathcal{R}_p$ to denote the $\ominus\boldsymbol{r}_p$-region graph for $\boldsymbol{r}_p$; $\ominus\mathcal{R}_p^O$ to denote its top (outer) regions; and $b_{\boldsymbol{r}_p}(\boldsymbol{x}_{\boldsymbol{\rho}})$ to denote the belief over region $\boldsymbol{\rho}$ in $\ominus\boldsymbol{r}_p$-region graph. For top-regions, the initial belief is equal to the basic messages obtained using Eqn (3).

Next we assign "counting numbers" to regions, in a way similar to CVM: top regions are assigned $\mathsf{cn}(\ominus\boldsymbol{r}_{p,q})=1$, and each sub-region $\boldsymbol{\rho}$ is assigned using

[1]This does not noticably affect the accuracy in our experiments. When using uniform cavity distributions, the results are identical.

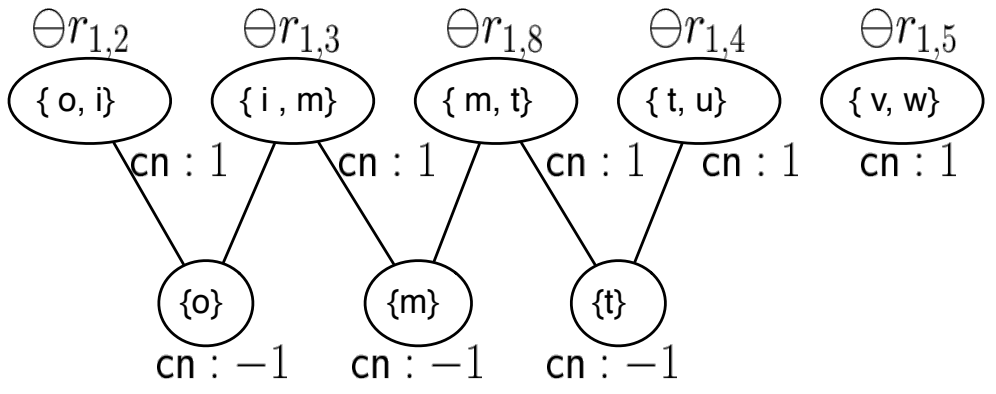

*Figure 2.* The $\ominus\boldsymbol{r}_1$-region-graph consisting of all the messages to $\boldsymbol{r}_1$. The variables in each region and its counting number are shown. The upward and downward messages are passed along the edges in this $\ominus\boldsymbol{r}_1$-region-graph.

the Möbius formula:

$$\mathsf{cn}(\boldsymbol{\rho}) \;:=\; 1-\sum\nolimits_{\boldsymbol{\rho}'\in\mathsf{A}(\boldsymbol{\rho})}\mathsf{cn}(\boldsymbol{\rho}')$$

where $\mathsf{A}(\boldsymbol{\rho})$ is the set of ancestors of $\boldsymbol{\rho}$.

We can now define the belief over cavity regions $\boldsymbol{r}_p$ as:

$$\hat{P}_{\boldsymbol{r}_p}(\boldsymbol{x}_{\oplus\boldsymbol{r}_p}) \propto \hat{P}_0^{\setminus\boldsymbol{r}_p}(\boldsymbol{x}_{\ominus\boldsymbol{r}_p})\psi_{N(\boldsymbol{r}_p)}(\boldsymbol{x}_{\oplus\boldsymbol{r}_p})\prod_{\boldsymbol{\rho}\in\ominus\mathcal{R}_p} b_{\boldsymbol{r}_p}(\boldsymbol{x}_{\boldsymbol{\rho}})^{\mathsf{cn}(\boldsymbol{\rho})} \quad (4)$$

This avoids any double-counting of variables, and reduces to Eqn (1) in the case of partitioning cavity regions.

To apply Eqn (4) effectively, we need to enforce marginal consistency of the intersection regions with their parents, which can be accomplished via message passing in a *downward pass*, Each region $\boldsymbol{\rho}'$ sends to each of its child $\boldsymbol{\rho}$, its marginal over the child's variables:

$$\mu_{\boldsymbol{\rho}'\to\boldsymbol{\rho}}(\boldsymbol{x}_{\boldsymbol{\rho}}) := \sum\nolimits_{\boldsymbol{x}_{\setminus\boldsymbol{\rho}}} b_{\boldsymbol{r}_p}(\boldsymbol{x}_{\boldsymbol{\rho}'})$$

Then set the belief over each child region to be the geometric average of the incoming messages:

$$b_{\boldsymbol{r}_p}(\boldsymbol{x}_{\boldsymbol{\rho}}) := \prod\nolimits_{\boldsymbol{\rho}'\in\mathsf{pr}(\boldsymbol{\rho})}\mu_{\boldsymbol{\rho}'\to\boldsymbol{\rho}}(\boldsymbol{x}_{\boldsymbol{\rho}})^{\frac{1}{|\mathsf{pr}(\boldsymbol{\rho})|}}$$

The downward pass updates the child regions in $\ominus\mathcal{R}_p\setminus\ominus\mathcal{R}_p^O$. We update the beliefs at the top regions using a modified version of Eqn (3): $b_{\boldsymbol{r}_p}(\boldsymbol{x}_{\ominus\boldsymbol{r}_{p,q}}) \propto$

$$\frac{\sum_{\boldsymbol{x}_{\setminus\ominus\boldsymbol{r}_{p,q}}}\hat{P}_{\boldsymbol{r}_q}(\boldsymbol{x}_{\oplus\boldsymbol{r}_q})\psi_{N(\boldsymbol{r}_q)\cap N(\boldsymbol{r}_p)}(\boldsymbol{x}_{\oplus\boldsymbol{r}_q})^{-1}}{\sum_{\boldsymbol{x}_{\setminus\ominus\boldsymbol{r}_{p,q}}}\hat{P}_{\boldsymbol{r}_p}(\boldsymbol{x}_{\oplus\boldsymbol{r}_p})\psi_{N(\boldsymbol{r}_p)\cap N(\boldsymbol{r}_q)}(\boldsymbol{x}_{\oplus\boldsymbol{r}_p})^{-1}} b_{\boldsymbol{r}_p}^{eff}(\boldsymbol{x}_{\ominus\boldsymbol{r}_{p,q}})^{\mathsf{cn}(\boldsymbol{\rho})}, \quad (5)$$

for all top regions $\ominus\boldsymbol{r}_{p,q}\in\ominus\mathcal{R}_p^O$.

Here $b_{\boldsymbol{r}_p}^{eff}(\boldsymbol{x}_{\ominus\boldsymbol{r}_{p,q}})$ is the effective old message over $\ominus\boldsymbol{r}_{p,q}$:

$$b_{\boldsymbol{r}_p}^{eff}(\boldsymbol{x}_{\ominus\boldsymbol{r}_{p,q}}) = \sum\nolimits_{\boldsymbol{x}\setminus\ominus\boldsymbol{r}_{p,q}}\prod\nolimits_{\boldsymbol{\rho}\in\ominus\mathcal{R}_p} b_{\boldsymbol{r}_p}(\boldsymbol{x}_{\boldsymbol{\rho}})$$

That is, in the update equation, we need the calculation of the new message to assume this value as the old message from $q$ to $p$. This marginalization is important because it allows the belief at the top region $b_{\boldsymbol{r}_p}(\boldsymbol{x}_{\ominus \boldsymbol{r}_{p,q}})$ to be influenced by the beliefs $b_{\boldsymbol{r}_p}(\boldsymbol{x}_{\boldsymbol{\rho}})$ of the sub-regions after a downward pass. It enforces marginal consistency between the top regions, and at convergence we have $b^{eff}_{\boldsymbol{r}_p}(\boldsymbol{x}_{\ominus \boldsymbol{r}_{p,q}}) = b_{\boldsymbol{r}_p}(\boldsymbol{x}_{\ominus \boldsymbol{r}_{p,q}})$. Notice also Eqn (5) is equivalent to the old update Eqn (3) in the partitioning case.

To calculate this marginalization more efficiently, GLC uses an *upward pass* in the $\ominus \boldsymbol{r}_p$-region-graph. Starting from the parents of the lowest regions, we define $b^{eff}_{\boldsymbol{r}_p}(\boldsymbol{x}_{\boldsymbol{\rho}})$ as:

$$b^{eff}_{\boldsymbol{r}_p}(\boldsymbol{x}_{\boldsymbol{\rho}'}) := b_{\boldsymbol{r}_p}(\boldsymbol{x}_{\boldsymbol{\rho}'}) \prod_{\boldsymbol{\rho} \in \mathsf{ch}(\boldsymbol{\rho}')} \frac{b^{eff}_{\boldsymbol{r}}(\boldsymbol{x}_{\boldsymbol{\rho}})}{\mu_{\boldsymbol{\rho}\to\boldsymbol{\rho}'}(\boldsymbol{x}_{\boldsymbol{\rho}})}$$

Returning to the example, the previous text provides a method to update $\hat{P}_{\boldsymbol{r}_1}(\boldsymbol{x}_{\oplus \boldsymbol{r}_1})$. GLC performs this for the remaining regions as well, and then iterates the entire process until convergence – *i.e.*, until the change in all distributions is less than a threshold.

## 4. Experiments

This section compares different variations of our method against LBP as well as CVM, LCBP and TreeEP (Minka & Qi, 2003) methods, each of which performs some kind of loop correction. For CVM, we use the double-loop algorithm of (Heskes, 2006), which is slower than GBP but has better convergence properties. All methods are applied without any damping. We stop each method after a maximum of 1E4 iterations or if the change in the probability distribution (or messages) is less than 1E-9. We report the time in seconds and the error for each method as the average of absolute error in single variable marginals – *i.e.*, $\sum_{\boldsymbol{x}_i, v} |\hat{P}(x_i = v) - P(x_i = v)|$. For each setting, we report the average results over 10 random instances of the problem. We experimented with grids, 3-regular random graphs, and the ALARM network as typical benchmark problems.[2]

Both LCBP and GLC can be used with a uniform initial cavity or with an initial cavity distribution estimated via clamping cavity variables. In the experiments, *full* and *uniform* refer to the kind of cavity distribution used. We use GLC to denote the partitioning case, and GLC+ when overlapping clusters of some form are used. For example, GLC+*(Loop4, full)* refers to a setting with full cavity that contains all overlapping loop clusters of length up to 4. If a factor does not appear in any loops, it forms its own cluster. The same form of clusters are used for CVM.

[2]The evaluations are based on implementation in *libdai* inference toolbox (Mooij, 2010).

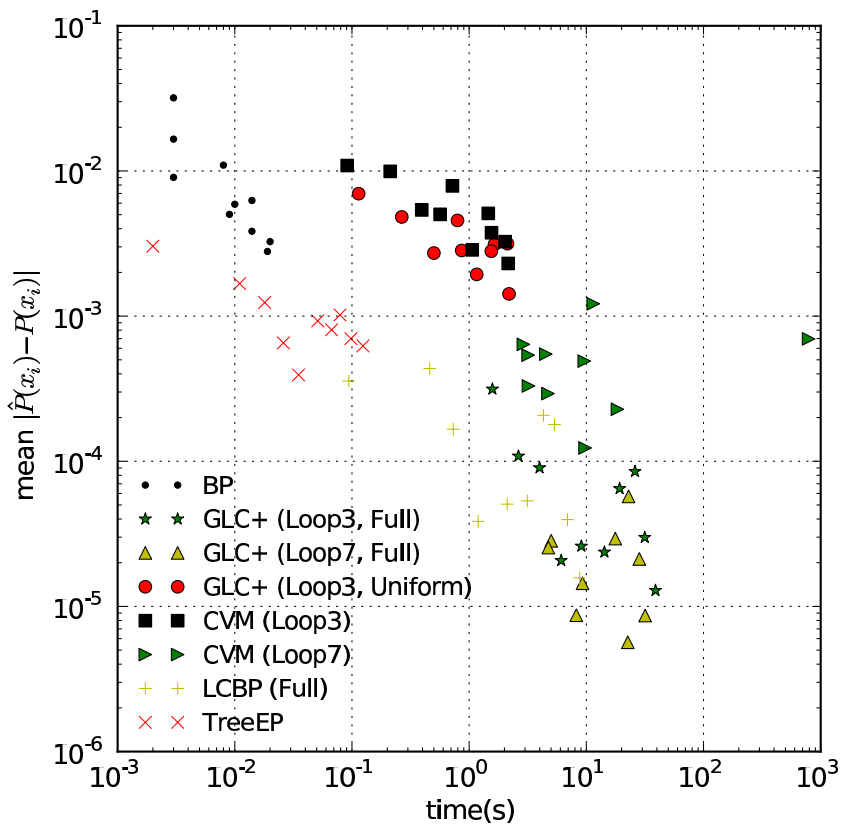

*Figure 4.* Time vs error for 3-regular Ising models with local field and interactions sampled from a standard normal. Each method in the graph has 10 points, each representing an Ising model of different size (10 to 100 variables).

### 4.1. Grids

We experimented with periodic Ising grids in which $x_i \in \{-1, +1\}$ is a binary variable and the probability distribution of a setting when $x_i$ and $x_j$ are connected in the graph is given by $P(\boldsymbol{x}) \propto \exp(\sum_i \theta_i x_i + \frac{1}{2}\sum_{i,j \in I} J_{i,j} x_i x_j)$ where $J_{i,j}$ controls variable interactions and $\theta_i$ defines a single node potential – *a.k.a.* a local field. In general, smaller local fields and larger variable interactions result in more difficult problems. We sampled local fields independently from $\mathcal{N}(0,1)$ and interactions from $\mathcal{N}(0,\beta^2)$. Figure 3(left) summarize the results for 6x6 grids for different values of $\beta$.

We also experimented with periodic grids of different sizes, generated by sampling all factor entries independently from $\mathcal{N}(0,1)$. Figure 3(middle) compares the computation time and error of different methods for grids of sizes that range from 4x4 to 10x10.

### 4.2. Regular Graphs

We generated two sets of experiments with random 3-regular graphs (all nodes have degree 3) over 40 variables. Here we used Ising model when both local fields and couplings are independently sampled from $\mathcal{N}(0,\beta^2)$. Figure 3(right) show the time and error for different values of $\beta$. Figure 4 shows time versus error for graph size between 10 to 100 nodes for $\beta = 1$. For larger $\beta$s, few instances did not converge within allocated number of iterations. The results are for cases in which all methods converged.

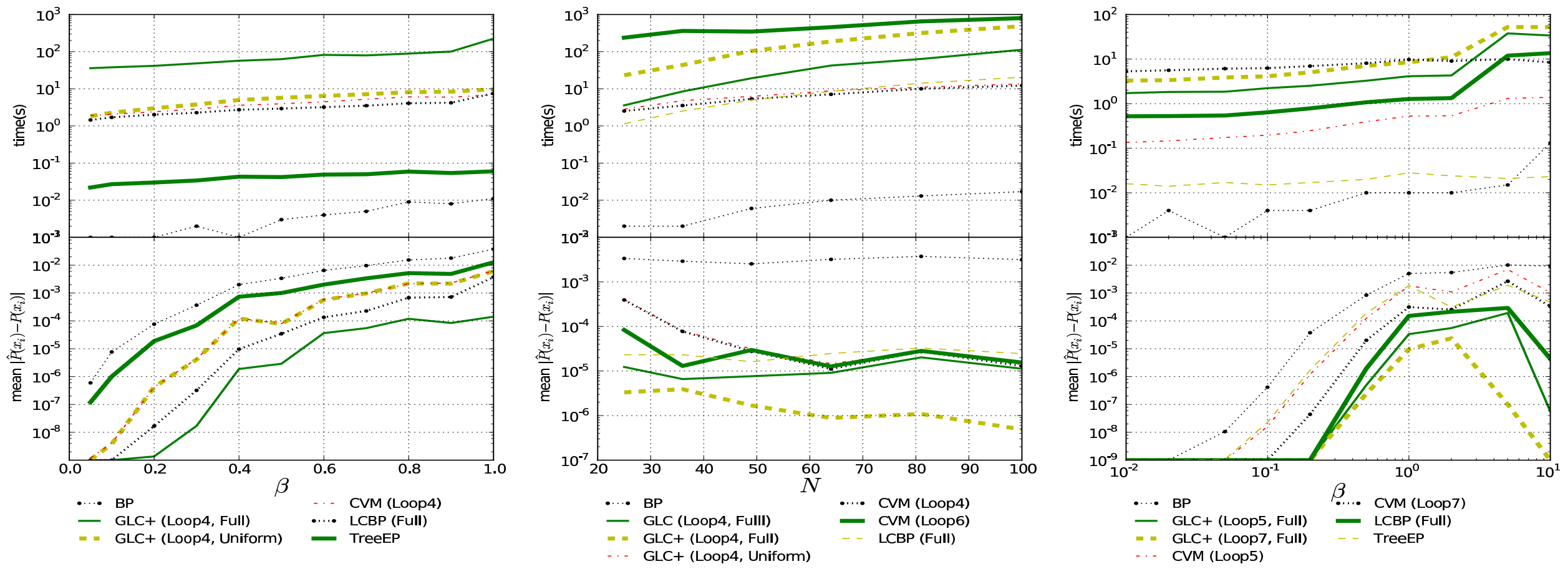

*Figure 3.* Average Run-time and accuracy for: **(Left)** 6x6 spinglass grids for different values of $\beta$. Variable interactions are sampled from $\mathcal{N}(0, \beta^2)$, local fields are sampled from $\mathcal{N}(0, 1)$. **(Middle)** various grid-sizes: [5x5, ..., 10x10]; Factors are sampled from $\mathcal{N}(0, 1)$. **(Right)** 3-regular Ising models with local field and interactions sampled from $\mathcal{N}(0, \beta^2)$.

*Table 1.* Performance of varoius methods on ALARM

| Method | Time(s) | Avg. Error |
|---|---|---|
| LBP | 3.00E-2 | 8.14E-3 |
| TreeEP | 1.00E-2 | 2.02E-1 |
| CVM (Loop3) | 5.80E-1 | 2.10E-3 |
| CVM (Loop4) | 7.47E+1 | 6.35E-3 |
| CVM (Loop5) | 1.22E+3 | 1.21E-2 |
| CVM (Loop6) | 5.30E+4 | 1.29E-2 |
| LCBP (Full) | 3.87E+1 | 1.07E-6 |
| GLC+ (Factor, Uniform) | 6.69E 0 | 3.26E-4 |
| GLC+ (Loop3, Uniform) | 6.71E 0 | 4.58E-4 |
| GLC+ (Loop4, Uniform) | 4.65E+1 | 3.35E-4 |
| GLC+ (Factor, Full) | 1.23E+3 | 1.00E-9 |
| GLC+ (Loop3, Full) | 1.36E+3 | 1.00E-9 |
| GLC+ (Loop4, Full) | 1.79E+3 | 1.00E-9 |

### 4.3. Alarm Network

ALARM is a Bayesian network with 37 variables and 37 factors. Variables are discrete, but not all are binary, and most factors have more than two variables. Table(1) compares the accuracy versus run-time of different methods. GLC with factor domains as regions – *i.e.*, $\boldsymbol{r}_p = I$ for $I \in \mathcal{F}$ – and all loopy clusters produces exact results up to the convergence threshold.

### 4.4. Discussions

These results show that GLC consistently provides more accurate results than both CVM and LCBP, although often at the cost of more computation time. They also suggest that one may not achieve this trade-off between time and accuracy simply by including larger loops in CVM regions. When used with uniform cavity, the performance of GLC (specifically GLC+) is similar to CVM, and GLC appears stable, which is lacking in general single-loop GBP implementations.

GLC's time complexity (when using full cavity, and using LBP to estimate the cavity distribution) is $\mathcal{O}(\tau M N |\mathcal{X}|^u + \lambda M |\mathcal{X}|^v))$, where $\lambda$ is the number of iterations of GLC, $\tau$ is the maximum number of iterations for LBP, $M$ is the number of clusters, $N$ is the number of variables, $u = \max_p |\ominus \boldsymbol{r}_p|$ and $v = \max_p |\oplus \boldsymbol{r}_p|$. Here the first term is the cost of estimating the cavity distributions and the second is the cost of exact inference on clusters. This makes GLC especially useful when regional Markov blankets are not too large.

## 5. Conclusions

We introduced GLC, an inference method that provide accurate inference by utilizing the loop correction schemes of both region-based and recent cavity-based methods. Experimental results on benchmarks support the claim that, for difficult problems, these schemes are complementary and our GLC can successfully exploit both. We also believe that our scheme motivates possible variations that can also deal with graphical models with large Markov blankets.

## 6. Acknowledgements

We thank the anonymous reviewers for their excellent detailed comments. This research was partly funded by NSERC, Alberta Innovates – Technology Futures (AICML) and Alberta Advanced Education and Technology.

## References

Aji, S and McEliece, R. The Generalized distributive law and free energy minimization. In *Allerton Conf*, 2001.

Braunstein, A., Mézard, M., and Zecchina, R. Survey prop-

agation: an algorithm for satisfiability. TR, 2002.

Heskes, T. Stable fixed points of loopy belief propagation are local minima of the Bethe free energy. In *NIPS*, 2003.

Heskes, T. Convexity arguments for efficient minimization of the Bethe and Kikuchi free energies. *JAIR*, 26, 2006.

Kikuchi, R. A theory of cooperative phenomena. *Phys. Rev.*, 81, 1951.

Koller, D. and Friedman, N. *Probabilistic Graphical Models: Principles and Techniques.* 2009.

Kschischang, F, Frey, B, and Loeliger, H. Factor graphs and the sum-product algorithm. *IEEE Info Theory*, 47, 1998.

Mézard, M. and Montanari, A. *Information, physics, and computation.* Oxford, 2009.

Mézard, M, Parisi, G, and Zecchina, R. Analytic and algorithmic solution of random satisfiability problems. *Science*, 2002.

Minka, T and Qi, Y. Tree-structured approximations by expectation propagation. In *NIPS*, 2003.

Montanari, A and Rizzo, T. How to compute loop corrections to the Bethe approximation. *J Statistical Mechanics*, 2005.

Mooij, J. *Understanding and Improving Belief Propagation.* PhD thesis, Radboud U, 2008.

Mooij, J. libDAI: A free and open source C++ library for discrete approximate inference in graphical models. *JMLR*, 2010.

Mooij, J and Kappen, H. Loop corrections for approximate inference on factor graphs. *JRML*, 2007.

Pearl, J. *Probabilistic reasoning in intelligent systems.* 1988.

Pelizzola, A. Cluster variation method in statistical physics and probabilistic graphical models. *J Physics A*, 2005.

Rizzo, T, Wemmenhove, B, and Kappen, H. On cavity approximations for graphical models. *J Physical Review*, 76(1), 2007.

Yedidia, J, Freeman, W, and Weiss, Y. Constructing free energy approximations and generalized belief propagation algorithms. *IEEE Info Theory*, 2005.

Yuille, A. CCCP algorithms to minimize the Bethe and Kikuchi free energies. *Neural Computation*, 2002.

# A. Appendix

We prove the equality of GLC to CVM, in the setting where each factor involves no more than 2 variables and the cavity distributions $\hat{P}^{\setminus \boldsymbol{r}}(\boldsymbol{x}_{\ominus \boldsymbol{r}})$ is uniform.[3]

Consider the following CVM region-graph:

- *internal* region ($\boldsymbol{R}^{int}_p$): it contains all the variables in $\boldsymbol{r}_p$, and factors that are internal to $\boldsymbol{r}_p$ – *i.e.*, $\{I \in \mathcal{F} \mid I \subseteq \boldsymbol{r}_p\}$.
- *bridge* region ($\boldsymbol{R}^{br}_{p,q}$): it contains all the variables and factors that connect $\boldsymbol{r}_p$ and $\boldsymbol{r}_q$ — *i.e.*, variables $\oplus \boldsymbol{r}_p \cap \oplus \boldsymbol{r}_q$ and factors $N(\boldsymbol{r}_p) \cap N(\boldsymbol{r}_q)$.
- *sub* region ($\boldsymbol{R}^{sub}_{p,q}$): the intersection of internal $\boldsymbol{R}^{int}_p$ and bridge $\boldsymbol{R}^{br}_{p,q}$. It contains only variables and no factors. (Note $\boldsymbol{R}^{sub}_{p,q} = \ominus \boldsymbol{r}_{q,p}$)

Note each internal and bridge region has a counting number of 1, while each subregion has a counting number of $-1$. Since we assume the cavity regions $\boldsymbol{r}_p$ form a partition and each factor contains no more than 2 variables, this region graph construction counts each variable and each factor exactly once.

We focus on the parent-to-child algorithm for GBP. For the specific region graph construction outlined, we have 2 types of messages: internal region to subregion message ($\mu^{is}_{q\to p}$ sent from $\boldsymbol{R}^{int}_q$ to $\boldsymbol{R}^{sub}_{q,p}$), and bridge region to subregion message ($\mu^{bs}_{q\to p}$ sent from $\boldsymbol{R}^{br}_{q,p}$ to $\boldsymbol{R}^{sub}_{q,p}$). Note that $\boldsymbol{R}^{sub}_{q,p}$ and $\boldsymbol{R}^{sub}_{p,q}$ are the intersection of $\boldsymbol{R}^{br}_{p,q}$ with $\boldsymbol{R}^{int}_q$ and $\boldsymbol{R}^{int}_p$ respectively. We use the notation $\mu$ to differentiate from messages $m$ used in GLC. Below we drop the arguments to make the equations more readable. The parent-to-child algorithm uses the following fixed-point equations:

$$
\begin{aligned}
\mu^{is}_{q\to p} &\propto \textstyle\sum_{\boldsymbol{x}\setminus \boldsymbol{R}^{sub}_{q,p}} \psi_{\boldsymbol{R}^{int}_q} \prod_{q' \in Nb(q), q' \neq p} \mu^{bs}_{q'\to q} \\
\mu^{bs}_{q\to p} &\propto \textstyle\sum_{\boldsymbol{x}\setminus \boldsymbol{R}^{sub}_{q,p}} \psi_{\boldsymbol{R}^{br}_{q,p}} \mu^{is}_{q\to p}
\end{aligned}
$$

Suppose GBP converges to a fixed point with messages $\mu^{is}_{q\to p}$ and $\mu^{bs}_{q\to p}$ satisfying the fixed point conditions above; we show that messages defined by $m_{q\to p} := \mu^{is}_{q\to p}$ are fixed points of update Eqn (3) – *i.e.*, satisfy the consistency condition of Eqn (2)

Assuming uniform initial cavity $\hat{P}^{\setminus \boldsymbol{r}}_0 = 1$, for LHS of Eqn (2), we have

$$
\begin{aligned}
&\textstyle\sum_{\boldsymbol{x}\setminus \ominus \boldsymbol{r}_{p,q}} \hat{P}_{\boldsymbol{r}_p}(\boldsymbol{x}_{\oplus \boldsymbol{r}_p}) \psi_{N(\boldsymbol{r}_p)\cap N(\boldsymbol{r}_q)}(\boldsymbol{x})^{-1} \\
\propto \quad & m_{q\to p} \textstyle\sum_{\boldsymbol{x}\setminus \ominus \boldsymbol{r}_{p,q}} \psi_{N(\boldsymbol{r}_p)\setminus N(\boldsymbol{r}_q)} \prod_{q' \in Nb(p), q' \neq q} m_{q'\to p} \\
\propto \quad & m_{q\to p} = \mu^{is}_{q\to p},
\end{aligned}
$$

as the domain of the expression inside the summation sign is disjoint from $\ominus \boldsymbol{r}_{p,q}$.

As for the RHS of Eqn (2) we have

$$
\begin{aligned}
& \sum_{\boldsymbol{x}\setminus \ominus \boldsymbol{r}_{p,q}} \hat{P}_{\boldsymbol{r}_q}(\boldsymbol{x}_{\oplus \boldsymbol{r}_q}) \psi_{N(\boldsymbol{r}_p)\cap N(\boldsymbol{r}_q)}(\boldsymbol{x})^{-1} \\
\propto & \sum_{\boldsymbol{x}\setminus \ominus \boldsymbol{r}_{p,q}} \psi_{N(\boldsymbol{r}_q)} \psi_{N(\boldsymbol{r}_p)\cap N(\boldsymbol{r}_q)}(\boldsymbol{x})^{-1} \prod_{q' \in Nb(q)} m_{q'\to q} \\
\propto & \sum_{\boldsymbol{x}\setminus \boldsymbol{R}^{sub}_{q,p}} (\psi_{\boldsymbol{R}^{int}_q} \prod_{q' \in Nb(q)} \psi_{\boldsymbol{R}^{br}_{q',q}}) \psi_{\boldsymbol{R}^{br}_{p,q}}(\boldsymbol{x})^{-1} \prod_{q' \in Nb(q)} \mu^{is}_{q'\to q} \\
\propto & \sum_{\boldsymbol{x}\setminus \boldsymbol{R}^{sub}_{q,p}} \psi_{\boldsymbol{R}^{int}_q} \prod_{q' \in Nb(q), q' \neq p} \psi_{\boldsymbol{R}^{br}_{q',q}} \mu^{is}_{q'\to q} \qquad (6) \\
\propto & \sum_{\boldsymbol{x}\setminus \boldsymbol{R}^{sub}_{q,p}} \psi_{\boldsymbol{R}^{int}_q} \prod_{q' \in Nb(q), q' \neq p} \sum_{\boldsymbol{x}\setminus \boldsymbol{R}^{sub}_{q',q}} \psi_{\boldsymbol{R}^{br}_{q',q}} \mu^{is}_{q'\to q} \qquad (7) \\
\propto & \sum_{\boldsymbol{x}\setminus \boldsymbol{R}^{sub}_{q,p}} \psi_{\boldsymbol{R}^{int}_q} \prod_{q' \in Nb(q), q' \neq p} \mu^{bs}_{q'\to q} \propto \mu^{is}_{q\to p}
\end{aligned}
$$

Removing $\mu^{is}_{p\to q}$ in line (6) is valid because, in the absence of $\psi_{\boldsymbol{R}^{br}_{p,q}}$, its domain is disjoint from the rest of the terms. Moving the summation inside the product in line (7) is valid because partitioning guarantees that product terms' domains have no overlap and they are also disjoint from $\psi_{\boldsymbol{R}^{int}_q}$.

Thus the LHS and RHS of Eqn (2) agrees and $m_{q\to p} := \mu^{is}_{q\to p}$ is a fixed point of GLC.

[3] To differentiate from GLC's cavity regions $\boldsymbol{r}$, we use the capital notation $\boldsymbol{R}$ to denote the corresponding region in the CVM region graph construction.